\documentclass[]{article}
\pdfoutput=1

\usepackage[alpha]{mdpn}

\usepackage[top=2cm, bottom=2cm, left = 1.5cm, right = 1cm,columnsep=20pt]{geometry}	
\usepackage{mathtools}
\usepackage{amsmath}
\usepackage{amsthm}
\usepackage{amssymb}
\usepackage[round]{natbib}						
\usepackage{hyperref}
\usepackage{xcolor}
\definecolor{dark-blue}{rgb}{0,0,0.7}
\definecolor{codegreen}{rgb}{0,0.6,0}
\definecolor{codegray}{rgb}{0.5,0.5,0.5}
\definecolor{codepurple}{rgb}{0.58,0,0.82}
\definecolor{backcolour}{rgb}{0.85,0.85,0.92}
\hypersetup{
    colorlinks, linkcolor={dark-blue},
    citecolor={dark-blue}, urlcolor={dark-blue}
}
\usepackage{mathrsfs} 

\usepackage{listings}
\lstdefinestyle{mystyle}{
    backgroundcolor=\color{backcolour},
    commentstyle=\color{codegreen},
    keywordstyle=\color{magenta},
    numberstyle=\tiny\color{codegray},
    stringstyle=\color{codepurple},
    basicstyle=\footnotesize\ttfamily,
    breakatwhitespace=false,
    breaklines=true,
    captionpos=b,
    keepspaces=true,
    numbers=left,
    numbersep=5pt,
    showspaces=false,
    showstringspaces=false,
    showtabs=false,
    tabsize=2
}
\author{Philip S. Thomas$^1$ and Billy Okal$^2$\\\\
$^1$Carnegie Mellon University, $^2$Albert-Ludwigs-Universit\"at Freiburg}

\title{A Notation for Markov Decision Processes}
\date{}

\begin{document}

\maketitle

\section{Introduction}
Many \textit{reinforcement learning} (RL) research papers contain paragraphs that define \textit{Markov decision processes} (MDPs). These paragraphs take up space that could otherwise be used to present more useful content. In this paper we specify a notation for MDPs that can be used by other papers. Declaring the use this notation using a single sentence can replace several paragraphs of notational specifications in other papers. Importantly, the notation that we define is a common foundation that appears in many RL papers, and is not meant to be a complete notation for an entire paper.

We refer to our notation as the \textit{Markov Decision Process Notation, version 1} or MDPNv1. It can be invoked in research papers with the sentence:
\begin{quote}
``We use the notational standard MDPNv1." 
\end{quote}
This sentence denotes that the notation specified in this document should be inserted at the current location. One challenge with this system is that any reasonably complete notation will define a large subset of the commonly used mathematical symbols, some of which an author may wish to use with a meaning other than that specified in MDPNv1. To overcome this problem, definitions that occur after the sentence invoking MDPNv1 can modify or overwrite the definitions in MDPNv1.

For example, an author may write ``We assume that the state and action sets are finite," which overrules MDPNv1's more general definition of the state and action sets, or ``Let $\mathcal A$ denote the set of all possible advantage functions," which overwrites the definition of $\mathcal A$ in MDPNv1 (where it is the set of possible actions). In general, MDPNv1 should serve as a notational foundation, which the author is free to build upon or remove from to best suit the needs of the paper.

This paper is \textit{not} an introduction to RL. It assumes that the reader is already familiar with the basic concepts of RL, as covered by \citet{SuttonBarto}. Also, we try to minimize the number of assumptions that we make. This means that authors using our notation will have to specify their own assumptions, rather than specify which of our assumptions must be removed.

Billy Okal has provided a style file for MDPNv1 at \url{https://github.com/makokal/MDPN}. Not only does this style file allow you to easily switch between the different notational variants defined below, but using it allows you to change the notation used in your paper by modifying the style file rather than by editing every equation individually.

\section{Discrete and Continuous Random Variables}
In general, the state, action, and reward at time $t$ can be discrete or continuous random variables, or even a mixture of both. A discrete random variable, $X$, that takes values in a set, $\mathcal X$, has a \textit{probability mass function} (PMF), $f:\mathcal X \to [0,1]$, such that $f(x) = \Pr(X=x)$ for all $x \in \mathcal X$. However, continuous random variables (and random variables that are a mixture of discrete and continuous) are not characterized by a PMF. Although measure theoretic probability offers a unified notation for discussing arbitrary random variables, its use is not commonplace in reinforcement learning literature, and so it may dilute the message of a paper and shrink a paper's audience.

We therefore introduce an abuse of notation into MDPNv1: notationally, we treat the state, action, and reward as though they are discrete random variables, even though they may not be. That is, our expressions are written using PMFs for distributions over states, actions, and rewards, even if they should technically be written using probability measures. The author of a paper using MDPNv1 should ensure that all claims carry over to states, actions, and rewards that are arbitrary random variables, or should explicitly restrict the states, actions, and rewards to be discrete random variables or continuous random variables that have density functions.

\section{Markov Decision Process Notation, Version 1 (MDPNv1)}
Let a \textit{Markov decision process} (MDP) be a tuple, $(\mathcal S, \mathcal A, \mathcal R, P, R, d_0, \gamma)$, where
\begin{enumerate}
	\item We use $t \in \mathbb N_{\geq 0}$ to denote the time step, where $\mathbb N_{\geq 0}$ denotes the natural numbers \textit{including zero}.
	
	\item $\mathcal S$ is the set of possible states that the agent can be in, and is called the \textit{state set}. The state of the environment at time $t$ is a random variable that we denote by $S_t$. We will typically use $s$ to denote an element of the state set.
	
	\item $\mathcal A$ is the set of possible actions that the agent can select between, and is called the \textit{action set}. The action chosen by the agent at time $t$ is a random variable that we denote by $A_t$. We will typically use $a$ to denote a specific element of the action set.
	
	\item $\mathcal R \subseteq \mathbb R \cup \{-\infty, \infty\}$ is the set of possible rewards that the agent can receive, and is called the \textit{reward set}. The reward provided to the agent at time $t$ is a random variable that we denote by $R_t$. We will typically use $r$ to denote an element of the reward set. Let $r_\text{min}$ and $r_\text{max}$ be the infimum and supremum of $\mathcal R$, respectively. 
	
	\item $P:\mathcal S \times \mathcal A \times \mathcal S \to [0,1]$ is called the \textit{transition function}. For all $(s,a,s',t) \in \mathcal S \times \mathcal A \times \mathcal S \times \mathbb N_{\geq 0}$, let $P(s,a,s')\coloneqq\Pr(S_{t+1}=s' | S_t=s, A_t=a)$.\footnote{Notice that we use $\coloneqq$ to denote ``is defined to be".} That is, $P$ characterizes the distribution over states at time $t+1$ given the state and action at time $t$. We introduce a \textit{Markov assumption}: the distribution over $S_{t+1}$ is independent of all prior events given $S_t$ and $A_t$. That is, the distribution over states at time $t+1$ is fully determined by the state and action at time $t$, and this distribution is characterized by $P$.

    We allow three alternate notations for $P$. First, let $P(s'|s,a)\coloneqq P(s,a,s')$. This form takes approximately the same amount of space, but makes it more clear that $P$ is a conditional distribution over the next state given the current state and action. Second, let $P_s^a(s')\coloneqq P(s,a,s')$. This notation moves terms into subscripts and superscripts in order to save some space. Third, let $P_{s,s'}^a\coloneqq P(s,a,s')$. This final form is particularly useful when space is limited. Although the author is allowed to select between the four notations for $P$, the use of $P$ should be consistent within each paper.

	\item $R$ is called the \textit{reward function}. For all $(s,a,s',t,r) \in \mathcal S \times \mathcal A \times \mathcal S \times \mathbb N_{\geq 0} \times \mathcal R$, let $R(s,a,s',r)\coloneqq \Pr(R_t=r | S_t=s,A_t=a,S_{t+1}=s')$. That is, $R$ characterizes the distribution over rewards at time $t$ given $S_t,A_t,$ and $S_{t+1}$. We introduce another Markov assumption: the distribution of $R_t$ is independent of all prior events given $S_t,A_t,$ and $S_{t+1}$. Also notice that the reward function, $R$, has no subscripts or superscripts, unlike the visually similar reward at time $t$, $R_t$.

    As with $P$, we allow for several alternate notations for $R$ that the author is free to select from. Let $R(r | s,a,s')\coloneqq R_{s,s'}^a(r) \coloneqq R_{s,s'}^{a,r} \coloneqq R(s,a,s',r)$.
    
	\item We call $d_0:\mathcal S \to [0,1]$ the \textit{initial state distribution}, since $d_0(s)\coloneqq \Pr(S_0=s)$ for all $s \in \mathcal S$.
	
	\item Let $\gamma \in [0,1]$ be the \textit{reward discount parameter}, which may be used to discount rewards based on how far in the future they occur.
\end{enumerate}

Let $\pi:\mathcal S \times \mathcal A \to [0,1]$ be called a \textit{policy}. A policy specifies the distribution over $A_t$ given $S_t$, i.e., $\pi(s,a)\coloneqq \Pr(A_t=a|S_t=s)$ for all $(s,a,t) \in \mathcal S \times \mathcal A \times \mathbb N_{\geq 0}$. All policies are assumed to be Markovian---the distribution of $A_t$ is independent of prior events given $S_t$. Let $\Pi$ be the set of all possible policies. If there exists a state, $s \in \mathcal S$, and two unique actions, $(a_1,a_2) \in \mathcal A^2$, where $a_1 \neq a_2$, and both $a_1$ and $a_2$ have non-zero probability in $s$, i.e., $\pi(s,a_1) > 0$ and $\pi(s,a_2) > 0$, then we refer to $\pi$ as a \textit{stochastic policy}, and we refer to it as a \textit{deterministic policy} otherwise. Let $\mu:\mathcal S \to \mathcal A$ be an alternate definition of a deterministic policy.  We allow several additional shorthands: $\pi(a|s)\coloneqq \pi_s(a)\coloneqq \pi_s^a\coloneqq \pi(s,a)$ and $\mu_s\coloneqq \mu(s)$.

We abuse notation and give $\pi$ a second definition. It should be clear from context which definition is intended. Let $\pi:\mathcal S \times \mathcal A \times \mathbb R^{{n_{\boldsymbol\theta}}} \to [0,1]$, where ${n_{\boldsymbol\theta}} \in \mathbb N_{\geq0}$. Let $\boldsymbol\theta \in \mathbb R^{{n_{\boldsymbol\theta}}}$ denote a ${n_{\boldsymbol\theta}}$-dimensional vector called the \textit{policy parameters}, and let $\pi(s,a,\boldsymbol\theta)\coloneqq \Pr(A_t=a | S_t=s, \boldsymbol\theta)$ for all $(s,a,\boldsymbol\theta,t) \in \mathcal S \times \mathcal A \times \mathbb R^{{n_{\boldsymbol\theta}}} \times \mathbb N_{\geq 0}$. We call this definition of $\pi$ a \textit{parameterized policy}. We allow several shorthands: $\pi(a|s,\boldsymbol\theta)\coloneqq \pi_{\boldsymbol\theta}(s,a) \coloneqq \pi_{\boldsymbol\theta}(a | s) \coloneqq \pi_{{\boldsymbol\theta}}^s(a) \coloneqq \pi_{{\boldsymbol\theta}}^{s,a}\coloneqq \pi(s,a,{\boldsymbol\theta})$. Similarly, $\mu:\mathcal S \times \mathbb R^{{n_{\boldsymbol\theta}}}\to \mathcal A$ is a parameterized deterministic policy, and $\mu_{\boldsymbol\theta}(s)\coloneqq \mu_{\boldsymbol\theta}^s \coloneqq \mu(s,{\boldsymbol\theta})$.

An \textit{episode} is one sequence of states, actions, and rewards, starting from $t=0$ and continuing indefinitely. An MDP may have a state, $\overset{\infty}{s} \in \mathcal S$, called the \textit{terminal absorbing state}. In the state $\overset{\infty}{s}$ only one action can be taken. Taking this action causes a transition back to $\overset{\infty}{s}$ and results in a reward of zero. Once the agent reaches $\overset{\infty}{s}$ the system has effectively terminated since there are no more decisions to be made or rewards to collect. If a state, $s \in \mathcal S$ always causes a transition to $\overset{\infty}{s}$ with a reward of zero, then we call $s$ a \textit{terminal state}. Let $L \in \mathbb N_{\geq 0} \cup \{\infty\}$ be the \textit{horizon} of the MDP, i.e., the smallest time step such that for all $\pi \in \Pi, \Pr(S_{L}=\overset{\infty}{s} | \pi)=1$, and $\infty$ if no such time step exists. 


\section{Discussion}

In this section we discuss some of the decisions that we made regarding notation. In general, we use calligraphic capital letters for sets, e.g., $\mathcal X$. Elements of sets are lowercase letters that are typically similar to the set they belong to, e.g., $x \in \mathcal X$. Random variables are denoted by capital letters, e.g., $X$, and their instantiations by lowercase letters, e.g., $x$. Vectors are bold lowercase letters, like $\mathbf x$.

Although we would have liked to use lowercase letters for real-valued functions, we use $P$ and $R$ to denote real-valued functions. This is for two different reasons. First, we use $P$ rather than $p$ because $p$ is a commonly used symbol that we would like to avoid defining (notice that we have not defined $x,y,i,j,f,g,p,$ or $q$, all of which are commonly used symbols). Second, we use $R$ because $r$ is already used to denote an element of $\mathcal R$, and to preserve alliteration we do not want to use a different letter. Although $R$ is visually similar to $R_t$, it is typically clear from context which is intended, even if the reader does not notice the subscript or lack thereof.

Sometimes the set of actions that can be selected by the agent changes depending on the state of the environment. We do not include this in our notation because it is rarely used in the literature. If the author wishes to include this additional structure in an MDP, then we recommend using $\mathscr{A}(s)$ to denote the set of actions that can be chosen in the state $s$. However, this is \textit{not} part of MDPNv1, and must be specified by the author.

Often MDPs are defined without explicitly defining the set of possible rewards, $\mathcal R$. We include $\mathcal R$ so that the author can write $\sum_{r \in \mathcal R} f(r)$ for some $f:\mathcal R \to \mathbb R$. This is useful because the two obvious choices for implicit definitions of $\mathcal R$ both have problems: $\sum_{r \in \mathbb R}$, while technically valid, may be confusing since the reals are typically integrated over, and $\sum_{r \in \mathbb Z}$ does not allow for rewards that are not integers.

Although there are many other terms that we could include in MDPNv1, we have decided to only define the terms necessary to define an MDP. This both makes it easier for the reader to remember which terms are defined by MDPNv1 and avoids including controversial definitions. Furthermore, it avoids limiting the setting to only the discounted or average-reward setting (we could define symbols for both settings, but this would be unnecessarily complex).

\section{\LaTeX\ Style File Usage}
\label{sec:usage}
In this section we demonstrate how to use the style file accompanying this text.

\begin{enumerate}
\item The package can be included using any of three options: {\tt alpha, beta, kappa}.
\begin{lstlisting}[language=Tex]
    % ...
    \usepackage[alpha]{mdpn}  % Most verbose
    %\usepackage[beta]{mdpn}  % Compressed
    %\usepackage[kappa]{mdpn}  % Most compressed
    % ...
\end{lstlisting}

\item You can use any of the defined commands in text as:
\begin{lstlisting}[language=Tex]
    % ...
    Some text $\command$, for example $\sset$ for state set
    % ...
\end{lstlisting}
    Some of the commands require a specific number of arguments that should be provided in the order indicated. For example {\tt \textbackslash T} requires three arguments: the current state $s$, current action $a$ and next state $s'$. So, {\tt \textbackslash T\{s\}\{a\}\{s'\} } will produce $\T{s}{a}{s'}$. 
\item Most of the commands allow usual modifications such as subscripts and superscripts. 
For example, {\tt \textbackslash pp} (which denotes a parametrised policy) can be modified to {\tt \textbackslash pp\_\{sub\}} to yield $\pp{s}{a}_{sub}$.
\end{enumerate}


\bibliographystyle{abbrvnat}	


\end{document}